\documentclass[10pt,twocolumn,letterpaper]{article}

\usepackage{wacv}
\usepackage{times}
\usepackage{epsfig}
\usepackage{graphicx}
\usepackage{amsmath}
\usepackage{amssymb}

\usepackage{multirow}
\usepackage{subfigure}
\usepackage{url}

\wacvfinalcopy % *** Uncomment this line for the final submission

\def\wacvPaperID{****} % *** Enter the wacv Paper ID here

\ifwacvfinal\fi
\begin{document}

%%%%%%%%% TITLE
\title{STELA: A Real-Time Scene Text Detector with Learned Anchor}

% Authors at the same institution
\author{Linjie Deng\textsuperscript{1}, Yanxiang Gong\textsuperscript{1}, Xinchen Lu\textsuperscript{1}, Yi Lin\textsuperscript{2}, Zheng Ma\textsuperscript{1}, Mei Xie\textsuperscript{1}\\
	\textsuperscript{1}School of Information and Communication Engineering, UESTC\\
	\textsuperscript{2}National Key Laboratory of Fundamental Science on Synthetic Vision, Sichuan University\\
}
% Authors at different institutions
%\author{First Author \\
%\and
%Second Author \\

\maketitle
\ifwacvfinal\thispagestyle{empty}\fi

%%%%%%%%% ABSTRACT
\begin{abstract}
To achieve high coverage of target boxes, a normal strategy of conventional one-stage anchor-based detectors is to utilize multiple priors at each spatial position, especially in scene text detection tasks. In this work, we present a simple and intuitive method for multi-oriented text detection where each location of feature maps only associates with one reference box. The idea is inspired from the two-stage R-CNN framework that can estimate the location of objects with any shape by using learned proposals. The aim of our method is to integrate this mechanism into a one-stage detector and employ the learned anchor which is obtained through a regression operation to replace the original one into the final predictions. Based on RetinaNet, our method achieves competitive performances on several public benchmarks with a totally real-time efficiency ($26.5fps$ at $800p$), which surpasses all of anchor-based scene text detectors. In addition, with less attention on anchor design, we believe our method is easy to be applied on other analogous detection tasks. The code will publicly available at \url{https://github.com/xhzdeng/stela}.
\end{abstract}

%%%%%%%%% BODY TEXT

\section{Introduction}
Text in scene usually conveys valuable semantic information. Thus, detecting text in natural images has recently attracted increasing attention in computer vision community cause perceiving information is a critical part of artificial general intelligence. It has been widely used in various applications such as multilingual translation, automotive assistance and image retrieval. Previous works \cite{SWT2010CVPR, RTTLR2012CVPR, FASTEXT2015ICCV, TF2015ICCV} have been dominated by sliding windows or connected component with hand-crafted feature, which divided the task into a sequence of distinct steps and utilized bottom-up strategy to search characters and words. Although these methods have shown their promising performances, they may be restricted to complex situations due to the diversity of text instances and undesirable image quality.

With the astonishing progress for object detection by exploring the powerful deep learning technology \cite{ALEXNET2012NIPS}, recent  methods take text as a specific object and extend the general object detection frameworks \cite{FASTERRCNN2015NIPS, YOLO2016CVPR, SSD2016ECCV} to hypothesize word or text locations. Those approaches can be divided into two major groups: two-stage proposal-driven and one-stage proposal-free method. Although two-stage framework \cite{DEEPTEXT2016ARXIV, RRPN2017ARXIV, RRCNN2017ARXIV} consistently achieves top accuracy on the public benchmarks \cite{ICDAR2013, ICDAR2015, ICDAR2017}, recent works \cite{TEXTBOX2017AAAI, DMPN2017CVPR, DEEPSPOTTER2017ICCV, SSTD2017ICCV} based on one-stage frameworks also demonstrate yielding faster text detectors with comparable accuracy. 

Unlike two-stage detector who can classify boxes at any position and shape by using learned proposals \cite{FASTERRCNN2015NIPS} and region pooling operation \cite{FASTRCNN2015ICCV}, one-stage detectors heavily rely on how densely the anchors cover the space of possible target locations \cite{RETINA2017ICCV}. A popular approach for achieving high coverage is to use multiple anchors to cover boxes of various scales and aspect ratios, especially in the tasks of scene text detection. TextBoxes++ \cite{TEXTBOX2017AAAI} was based on SSD \cite{SSD2016ECCV} and defined 7 specific aspect ratios (including 1, 2, 3, 5, 1/2, 1/3 and 1/5) for default boxes on each location of feature maps. In order to achieve multi-oriented text detection, DMPNet \cite{DMPN2017CVPR} added several rotated anchors, for a total of 12 (6 regular and 6 inclined) to find the best match to arbitrary-oriented text instance. Instead of choosing priors by hand, DeepTextSpotter \cite{DEEPSPOTTER2017ICCV} followed YOLOv2 \cite{YOLOV22017CVPR} runs k-means clustering $(k=14)$ on the training set bounding boxes to automatically find suitable priors.

Given the anchor design of the above detectors, a natural question to ask is: could we decrease the number of anchors and maintain similar accuracy? This changing will bring twofold benefit: reducing manual attention on anchors and improving efficiency at inference stage. First, the shapes and scales of anchors has to be predefined for different tasks, and this must be careful because a wrong design may harm the performance of detection \cite{GUIDED2019ARXIV}. Second, most anchors correspond to false candidates which are irrelevant to the targets, and meanwhile a large number of anchors can lead to significant computational cost when the network involves heavy heads. Besides, although not mentioned in many papers, the anchor generation usually needs to cost a certain amount of time. 

\begin{figure*}[htb]
	\begin{center}
		\subfigure[ICDAR 2013]{
			\label{fig:recall_a} 
			\includegraphics[height=6.0cm]{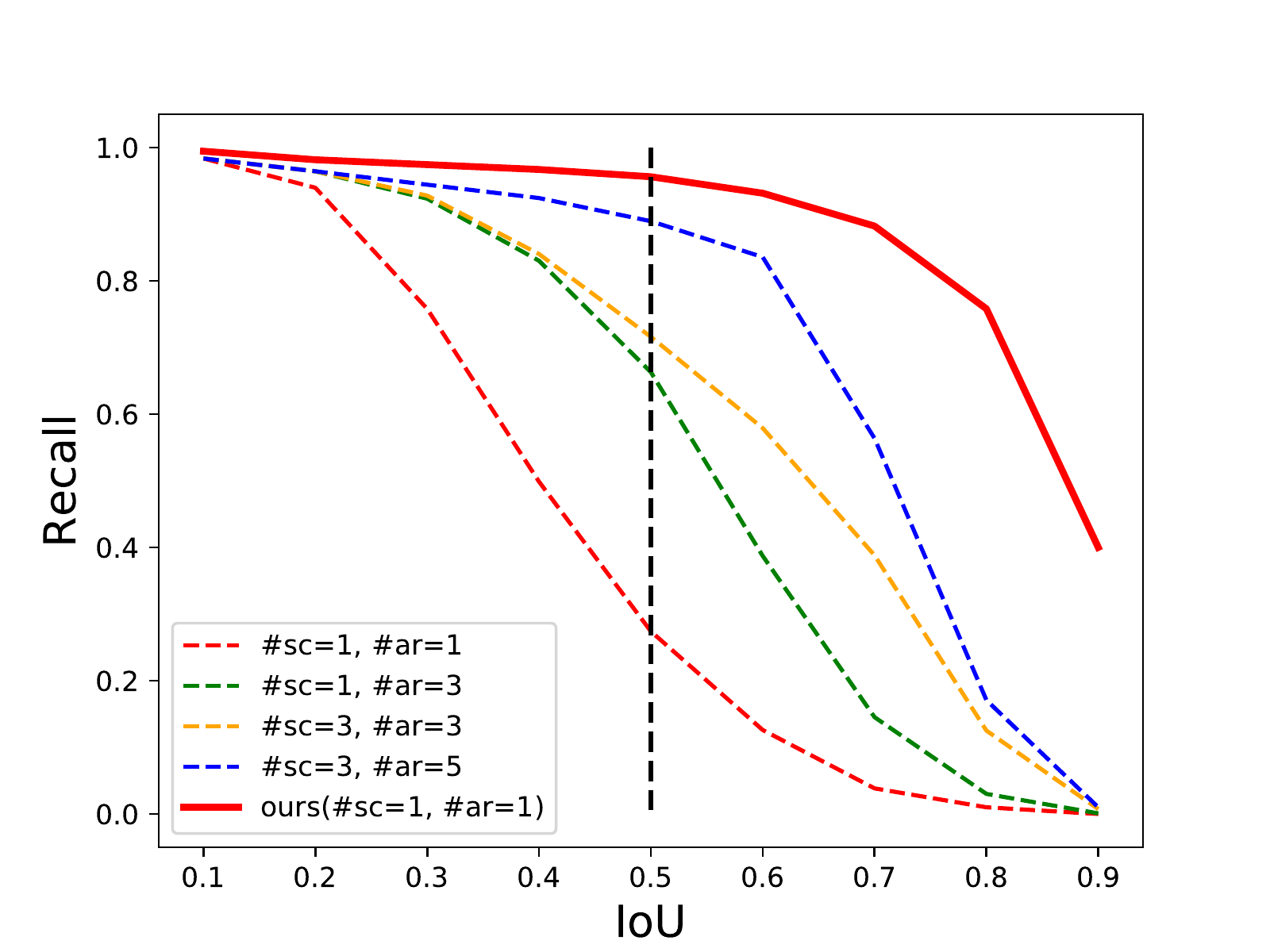}
		}
		\subfigure[ICDAR 2015]{
			\label{fig:recall_b} 
			\includegraphics[height=6.0cm]{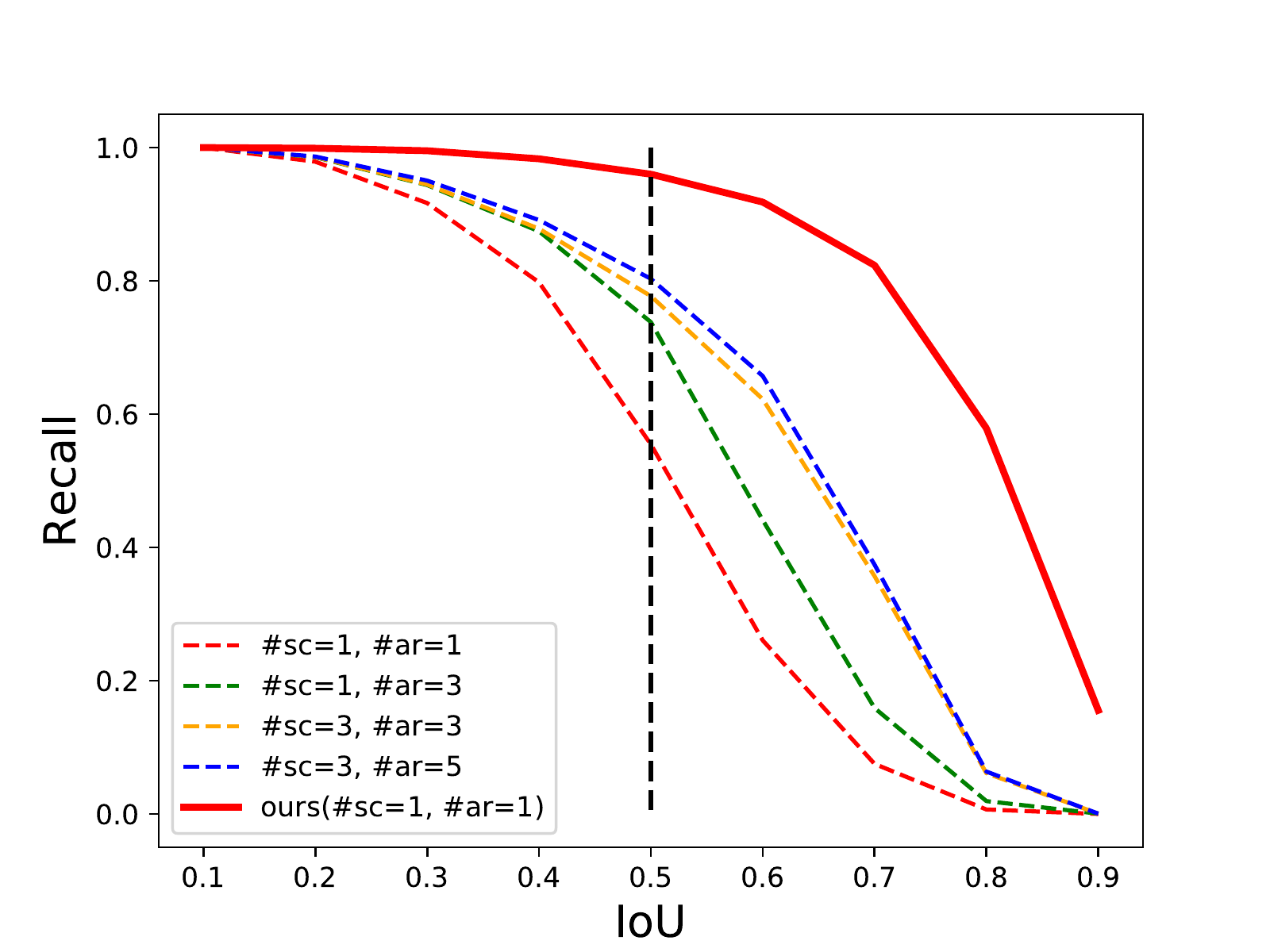}
		}
	\end{center}
	\caption{\textbf{The recall rates of different anchor designs on ICDAR 2013(a) and 2015(b).} "\#{sc}" means number of scales, "\#{ar}" means number of aspect ratios. The black line represents the IoU threshold ($0.5$) which is usually used in training stage to discriminate foreground and background. The other dashed lines with colors represent different anchor designs with various number of scales and aspect ratios. The solid line represent the recall rates of learned anchor which proposed in this work. The performances of each design will be shown in Section.\ref{sec:ablation}.}
	\label{fig:recall}
\end{figure*}

Being attracted to simple network architecture and high computational efficiency, in this work, we investigate the issue of anchor design within one-stage detector which we mentioned above for multi-oriented text detection. In one-stage methods, the optimization target in training and the prediction reference in testing are both based on the coverage between original anchors and target boxes. Then, the quality of those prior boxes has a critical impact on the performances of a detector. Normally, as the number of anchors increases, the coverage of targets increases, but it will still be saturated in some situations, as shown in Figure.\ref{fig:recall_b}. Therefore, we need to find a better way to choose priors that make it easier for the network to learn to predict better detection. Inspired from the learned proposal mechanism \cite{FASTERRCNN2015NIPS} in the two-stage R-CNN framework, we intend to utilize the learned anchor which is obtained through a regression operation to replace the original one into the final predictions. It is worth noting that unlike region proposal network (RPN) in two-stage detector which can reduce the number of possible locations down to one or two thousands,  we still maintain the original quantity of anchors and keep the rest parts of one-stage detector's architecture. To validate its effectiveness, we adopt the state-of-the-art RetinaNet \cite{RETINA2017ICCV} as our baseline model and present a simple and intuitive text detector named STELA ({\bf S}cene {\bf TE}xt Detector with {\bf L}earned {\bf A}nchor), in which each location of feature maps only associates with one anchor. Following the standard evaluation protocols in each benchmark, our method achieves comparable performances with an F-measure 0.887 on ICDAR 2013 \cite{ICDAR2013}, 0.833 on ICDAR 2015 \cite{ICDAR2015} and 0.715 on ICDAR 2017 MLT \cite{ICDAR2017}. Besides, our method is a totally real-time scene text detector with $26.5fps$ at $800p$, which surpasses all of anchor-based methods. At last, with less attention on anchor design, we believe our method is easy to be applied on other analogous detection tasks. Also, all of our training and testing code will open source soon.

%-------------------------------------------------------------------------
\section{Methodology}
\subsection{One-Stage Object Detection}
\begin{figure*}[htb]
	\begin{center}
		\subfigure[One-Stage Detector]{
			\label{fig:pipeline_a} 
			\includegraphics[height=3.25cm]{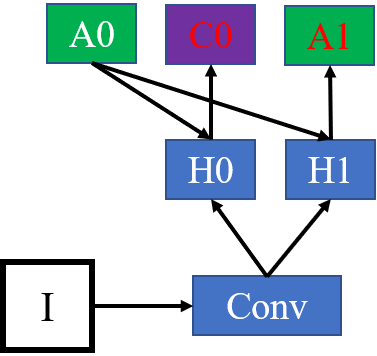}
		}
		\hspace{0.3cm}
		\subfigure[Faster R-CNN]{
			\label{fig:pipeline_b} 
			\includegraphics[height=3.25cm]{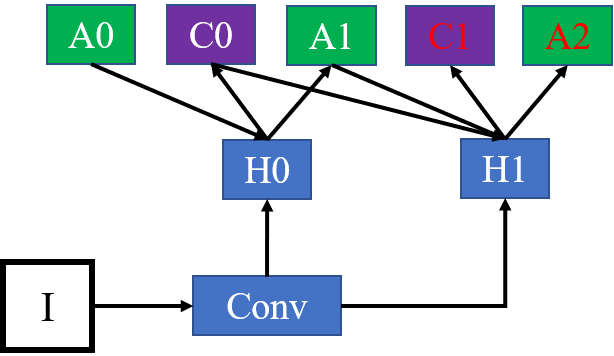}
		}
		\hspace{0.3cm}
		\subfigure[Ours]{
			\label{fig:pipeline_c} 
			\includegraphics[height=3.25cm]{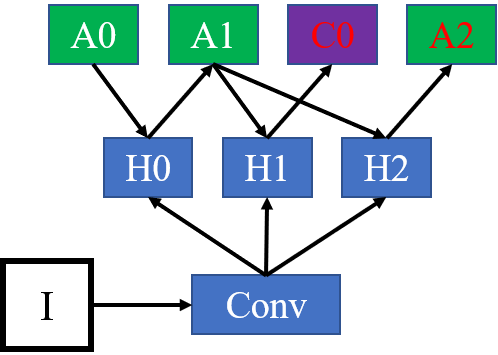}
		}
	\end{center}
	\caption{\textbf{The architectures of different frameworks.} "I" input image, "Conv" backbone convolutions, "H" network head, "C" classification. "A0" are the original anchors in all architectures, "A1" in (b) and (c) represent the selected proposals and learned anchors respectively. "A" and "C" in red color represent the final outputs of each detector.}
	\label{fig:pipelines}
\end{figure*}

In this section, we first review the one-stage detection pipeline. OverFeat \cite{OVERFEAT2013ARXIV} is one of the first modern one-stage object detector based on deep neural networks. More recent SSD \cite{SSD2016ECCV} and YOLOv2 \cite{YOLOV22017CVPR} have renewed interest in one-stage methods. The key idea of them is to associate a set of pre-defined anchors which are centered at each location of feature maps and make final predictions based on those reference boxes \cite{CONRETINA2019ARXIV}. As shown in Figure.\ref{fig:pipeline_a}, it basically contains a backbone network for feature extraction over the entire image and two parallel sub-networks following, one for predicting the probability distribution over multiple categories of each anchor and another for regression the offset from each positive candidate to a nearby ground-truth box, if one exists. 

Comparing with two-stage R-CNN (Figure.\ref{fig:pipeline_b}) methods, one-stage detectors skip the region proposal generation step and gives final predictions (classification and regression) based on original anchors directly. However, its detection accuracy is usually behind that of two-stage approaches, one of the main reasons is they must process a much larger set of candidate object locations regularly sampled across an image. The extreme foreground-background class imbalance problem will encounter during training phase and hamper the resulting performance. More recently, RetinaNet \cite{RETINA2017ICCV} proposed focal loss (FL) to address the class imbalance problem that one-stage detectors is able to match the accuracy of existing two-stage ones. The focal loss is modified from standard cross entropy (CE) loss:

\begin{gather}
CE(p, y) = \left\{
\begin{array}{ll}
{-log(p)} & \text{if}\ y=1 \, \\
{-log(1-p)} & \text{otherwise} \,
\end{array}
\right.
\label{eq:celoss}
\end{gather}

\noindent In the above $y=1$ specifies the ground-truth class and $p\in[0,1]$ is the probability. Normally, we define $p_t$:

\begin{gather}
p_t = \left\{
\begin{array}{ll}
{p} & \text{if}\ y=1 \, \\
{1-p} & \text{otherwise} \,
\end{array}
\right.
\label{eq:prob}
\end{gather}

\noindent and rewrite $CE(p,y)=-log(p_t)$. Then, the classification loss is defined as:

\begin{gather}
L_{cls} = FL(p_t) = -\alpha_{t}(1-p_t)^{\gamma}log(p_t)
\label{eq:focalloss}
\end{gather}

\noindent where $\alpha_{t}$ is a $balanced$ weighting factor and $\gamma$ is a $focusing$ parameter. It applies a modulating term to the cross entropy loss in order to focus learning on hard examples and down-weight the numerous easy negatives. In our implementation, we follow the original focal loss that set $\alpha_{t}=0.25$ and $\gamma=2.0$.

\subsection{Rotated Bounding Box Regression}
\label{sec:regression}
As depicted in \cite{CRPN2019NEUCO}, using rectangular bounding boxes to localize multi-oriented text may result in redundant background noise and unnecessary overlap. Thus, we adopt rotated rectangular boxes to match arbitrary-oriented text instances. Each bounding box $b$ is represented by a five tuple $b=(x,y,w,h,\theta)$,  where $x,y$ are the center point, $w,h$ are width and height, $\theta$ is the angle to horizontal. The task of the regression operation is to predict the distance of each item from a positive anchor to the nearby ground-truth. Normally, to encourage a regression invariant to scale and location, the distance vector $\Delta=(\delta_x, \delta_y, \delta_w, \delta_h, \delta_\theta)$ is defined by:

\begin{gather}
\delta_x = (g_x - b_x) / b_w, \quad \delta_y = (g_y - b_y) / b_h \\
\delta_w = log(g_w/b_w), \quad \delta_h = log(g_h/b_h) \\ 
\delta_\theta = tan(g_\theta) - tan(b_\theta)
\label{eq:delta}
\end{gather}

\noindent where $b$ and $g$ represent a bounding box and its target ground-truth respectively. The regression task loss $L_{loc}$ is calculated by regression target $\Delta_t$ and predicted tuple $\Delta_p$

\begin{gather}
L_{loc}=smooth_{L_1}(\Delta_t-\Delta_p)
\label{eq:regress}
\end{gather}

\noindent where $smooth_{L_1}$ is a robust $L_1$ loss defined in \cite{FASTRCNN2015ICCV}. Usually, for improving the effectiveness of multi-task learning, $\Delta$ is normalized by its mean and variance. In our experiments, the mean is set to (0, 0, 0, 0, 0) and the variance is set to (0.1, 0.1, 0.2, 0.2, 0.1).

\subsection{Learned Anchor}

\begin{figure*}[htb]
	\begin{center}
		\includegraphics[height=3.8cm]{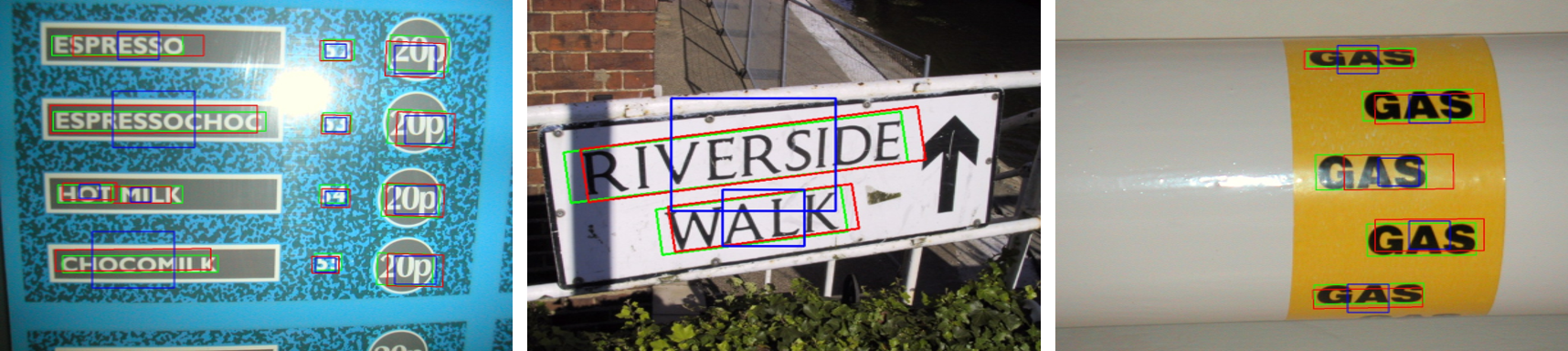}
	\end{center}
	\caption{\textbf{The view of different boxes}. The blue, red and green boxes represent original anchor, learned anchor and final output box respectively. It is worth noting that the learned anchor (red) shares the same central point with the original one (blue). Viewing digitally with zooming is recommended.}
	\label{fig:anchors}
\end{figure*}

Normally, the detector needs to search the true positives from thousands anchors and adjusts the shapes and locations to make them tighter on the targets. It is difficult to determine if a bounding box $b$ is a positive candidate cause it usually includes an object and some amount of the background. In practice, this is solved by the IoU metric between box $b$ and most nearby ground-truth $g$. Commonly, the threshold $\mu$ is a constant set to 0.5. If the IoU is above the threshold $\mu$, bounding box $b$ is considered to be an example of positive.

\begin{align}
y = \left\{
\begin{array}{ll}
1,  & \ IoU(b, g) \ge \mu \, \\
0,  & \text{otherwise} \,
\end{array}
\right.
\label{eq:class_label}
\end{align}

\noindent Also, $y=1$ specifies the ground-truth class. It is worth noting that conventional IoU based on rectangular boxes is unsatisfactory for our task, thus we modify it to compute the overlaps for rotated rectangles. Given all this, the optimization target in training is determined by the overlaps between original anchors with ground-truth boxes.

However, the original anchors with fixed scales and aspect ratios which are pre-defined manually may not be the optimal designs. Compared with one-stage detector, we argue that the most important part of proposal scheme in two-stage is that the selected proposals are chosen by learning. That makes two-stage method able to reduce the search space of targets, and meanwhile optimize the quality of candidates. Inspired by this, we intend to integrate this mechanism into the one-stage detectors. We simply add an extra regression branch for anchor refining and utilize learned one into the final classification and regression, as shown in Figure.\ref{fig:pipeline_c}. 

Especially, the regression targets of learned anchor is not arbitrary. As refer in \cite{GUIDED2019ARXIV}, one of general rules for a reasonable anchor design is alignment. To use convolutional features as anchor representations, the center of an anchor need to be well aligned with feature map pixels. Towards this end, we only regress the offsets within $\Delta^{'}=(\delta_w^{'}, \delta_h^{'}, \delta_\theta^{'})$, and this will keep anchors still align with feature map, as depicted in Figure.\ref{fig:anchors}. Following regression task, the anchor refining loss $L_{ref}$ is defined as:

\begin{gather}
L_{ref}=smooth_{L_1}(\Delta^{'}_t-\Delta^{'}_p)
\label{eq:refine}
\end{gather}

\noindent Unlike two-stage R-CNN method filter anchors with an objectness score, we only adjust the shape of each anchor and keep the quantity of anchors here. Evaluated on public benchmarks, we find comparing with original ones, the coverage of targets will be given a huge enhancement after anchor refining stage, as shown in Figure.\ref{fig:recall}.

\subsection{Network Architecture}
For the trade-off between efficiency and accuracy, all of our experiments are implemented on RetinaNet \cite{RETINA2017ICCV} with ResNet-50 \cite{RESNET2016CVPR} as backbone, though other networks are still applicable. We also adopt the Feature Pyramid Network (FPN) from \cite{FPN2017CVPR} to construct a rich, multi-scale feature pyramid from a single resolution input image. The FPN consists of levels $P_3$ to $P_7$ feature maps, and the corresponding base anchor sizes from $16^2$ to $256^2$ for detecting small text instances ($32^2$ to $512^2$ in source implementation). In original RetinaNet, the two sub-networks (heads) are deeper with 5 convolutional layers. For improving the running speed, we decrease the number of layers from 5 to 2 for streamlining the heads. This may result in a slight accuracy loss, but will give us a real-time text detector in return. Based on the above definitions, the model is trained to simultaneously minimize the losses on anchor refining, final regression and classification. Overall, the loss function is a weighted sum of three losses

\begin{gather}
L = \lambda_{ref}L_{ref} + \lambda_{loc}L_{loc} + \lambda_{cls}L_{cls}
\label{eq:loss_all}
\end{gather}

\noindent where $\lambda_{ref}$, $\lambda_{loc}$,  $\lambda_{cls}$ are user constants indicating the relative strength of each component defined above. In order to keep the balance of different loss types, we set them to 0.5, 0.5, 1.0 respectively.

%-------------------------------------------------------------------------
\section{Experiments}
\subsection{Implementation Details}
The backbone of network is initialized by the model trained on ImageNet \cite{IMAGENET2015IJCV} for classification task and other layers are initialized by following \cite{RETINA2017ICCV}. The network is trained with Adam \cite{ADAM2014ARXIV} optimizer. Restricted by the hardware, the batch size is set to $4$ and the initial learning rate is set to $10^{-4}$. We randomly pick up 100,000 images from SynthText \cite{FCRN2016CVPR} to pretrain the network for 5 epochs, and collect real data from ICDAR 2013 \cite{ICDAR2013}, 2015 \cite{ICDAR2015} and 2017 \cite{ICDAR2017} to finetune a final model for 25 epochs. The learning rate is decayed to $10^{-5}$ after 15 epochs of finetuning. We use the multi-scale training scheme that randomly resize the input size between 480 and 800. Random flipping is also used for data augmentation. 

Specially, in order to capture more regression target candidates, we set the IoU threshold to 0.3 in anchor refining training. In the inference stage, a confidence threshold with 0.3 and a non-maximum suppression threshold with 0.3 are applied to yield the final outputs. The proposed method is implemented by using PyTorch \footnote{\url{https://pytorch.org/}} and all experiments are carried out on a standard PC with Intel i7-6800k and a single NVIDIA TITAN Xp.

\subsection{Ablation Study}
\label{sec:ablation}

To investigate the effectiveness of our method, we conduct several ablation studies. Each model is evaluated on ICDAR 2013 \cite{ICDAR2013} and 2015 \cite{ICDAR2015} benchmarks.

{\bf Anchor Design:} We first investigate the impacts of different anchor designs on performances, including accuracy and efficiency. The baseline models are directly extended from RetinaNet by simply changing the regression strategy introduced in Section.\ref{sec:regression}. The aspect ratios of anchors on single location of feature maps are simply selected from \{0.25, 0.5, 1, 2, 4\}. Also, the scales are chosen from ($2^{k/3}, k<3$). As shown in Table.\ref{tab:baseline}, with the anchor increasing, the F-measure improved from 0.447 to 0.863 on ICDAR 2013, but no significant improvement (0.621 to 0.753) on ICDAR 2015. Analyzing from Figure.\ref{fig:recall}, we argue that the coverage of targets gets saturated on ICDAR 2015, but not on ICDAR 2013. That proves once again the most important design factor in a one-stage detector is how densely it covers the space of target boxes. 

Attaching anchor refining operation on the first baseline, the resulting model obtains a huge improvement (shown in Figure.\ref{fig:recall}) in recall rates on both benchmarks, with great progresses on accuracy from 0.447 to 0.887 on ICDAR 2013 and 0.621 to 0.833 on ICDAR 2015. This strongly demonstrates the effectiveness of our approach. In addition, we also assess other baselines with anchor refining, there is only slower running speed, but no obvious improvement.

\begin{table}[htb]
	\begin{center}
		\begin{tabular}{ccc|c|c|r}
			\#{sc} & \#{ar} & {la}       & ICDAR 2013  & ICDAR 2015  & FPS        \\
			\hline
			1      & 1      &            & 0.447       & 0.621       & {\bf 27.3} \\
			1      & 3      &            & 0.789       & 0.742       & 26.3       \\
			3      & 3      &            & 0.788       & 0.742       & 24.1       \\
			3      & 5      &            & 0.863       & 0.753       & 22.3       \\
			\hline
			1      & 1      & \checkmark & {\bf 0.887} & {\bf 0.833} & 26.5       \\
		\end{tabular}
	\end{center}
	\caption{{\bf The impact of the different anchor designs.} "\#{sc}" number of scales, "\#{ar}" number of aspect ratios. "la" means learned anchor. All input images are resize to 800 pixels.}
	\label{tab:baseline}
\end{table}

{\bf Number of Stages:} Like Cascade R-CNN \cite{CASCADE2018CVPR}, we add more stages of anchor refining to compare the influences. We also increase the IoU threshold of each refining stage by following Cascade R-CNN. The results are summarized in Table.\ref{tab:stages}. Increasing more refining stages will not lead to significant improvement, or even accuracy decrease. Besides, adding the number of stages will affect the running speed. Therefore, one refining stage is the best choice for our method. 

\begin{table}[htb]
	
	\begin{center}
		\begin{tabular}{c|c|c|r}
			\#{stages}       & ICDAR 2013  & ICDAR 2015  & FPS        \\
			\hline
			1                & 0.887       & {\bf 0.833} & {\bf 26.5} \\
			2                & {\bf 0.889} & 0.830       & 25.1       \\
			3                & 0.879       & 0.827       & 23.7       \\
		\end{tabular}
	\end{center}
	\caption{{\bf The impact of the number of stages.} "\#{stage}" means the number of anchor refining stages.}
	\label{tab:stages}
\end{table}

\subsection{Comparison to State of the Art}
\begin{figure*}[htb]
	\begin{center}
		\includegraphics[height=11.0cm]{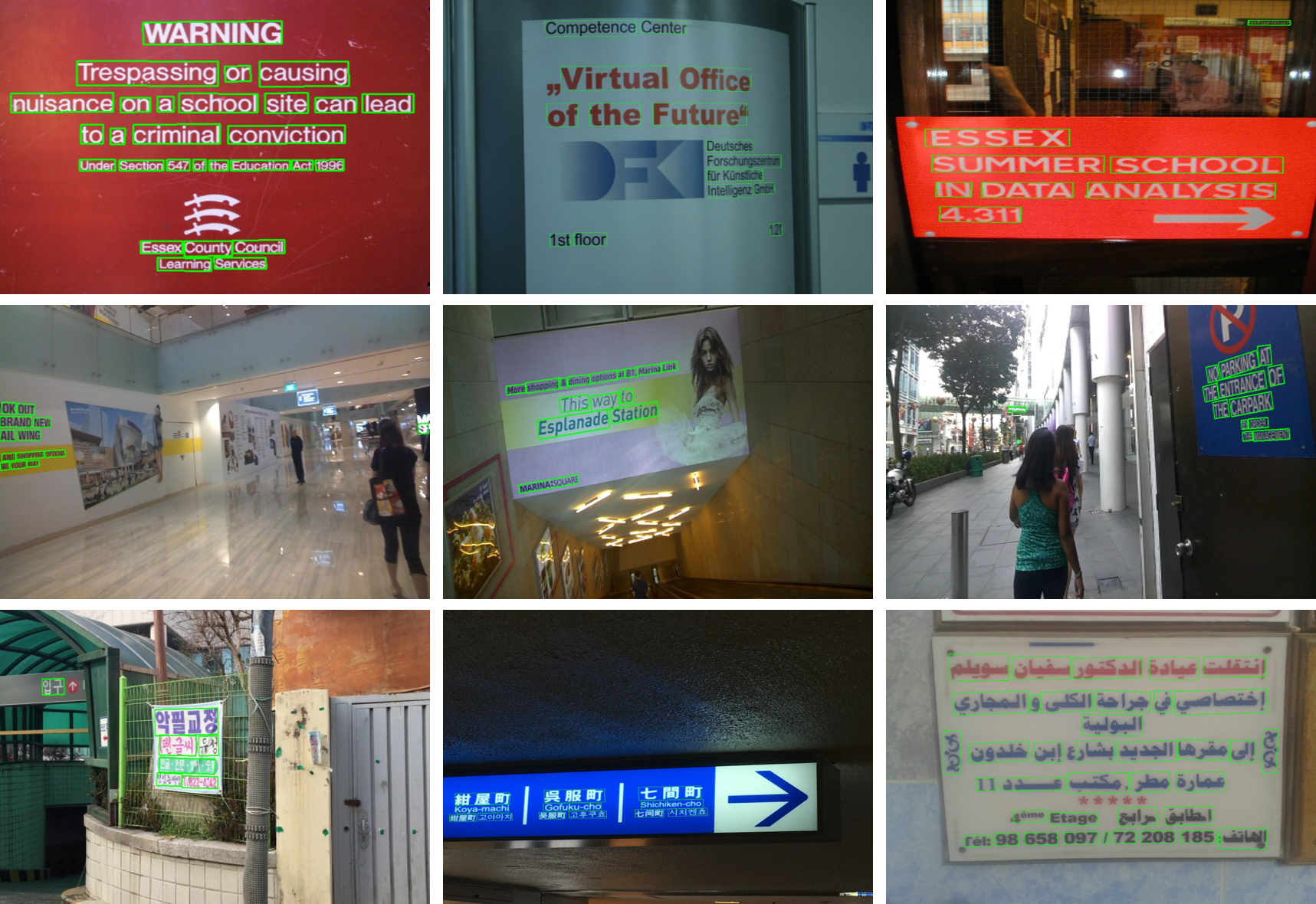}
	\end{center}
	\caption{\textbf{Selected results from the public benchmarks.} Viewing digitally with zooming is recommended.}
	\label{fig:results}
\end{figure*}

We evaluate our method on several public benchmarks and compare to recent state-of-the-art methods. Figure.\ref{fig:results} shows some detection results from each dataset. 

{\bf ICDAR 2013} \cite{ICDAR2013} dataset consists of 229 training and 233 testing images which were captured by user explicitly detecting the focus of the camera on the text content of interest. It is a standard benchmark for evaluating horizontal or nearly horizontal text detection. In this benchmark, we set the scale of input images to 800 for single-scale testing. We also evaluate on multi-scale testing which the scales are set to 320, 480, 640 and 800. As depicted in Table.~\ref{tab:icdar2013}, the proposed method outperforms all anchor based methods including DeepText \cite{DEEPTEXT2016ARXIV}, FCRN \cite{FCRN2016CVPR} and CTPN \cite{CTPN2016ECCV}, which are mainly designed for nearly horizontal text detection. For single-scale testing, our method achieves a totally real-time running speed at 26.5 fps. Even the multi-scale, our method runs at a speed of 10.5 fps. Compared with recent methods \cite{FOTS2018CVPR, RRD2018CVPR, LYU2018CVPR, CRPN2019NEUCO}, our method is comparable with accuracy and efficiency.

\begin{table*}[htb]
	\begin{center}
		\begin{tabular}{c|c|c|c|c|c|c|r} 
			\multirow{2}*{} & \multicolumn{3}{c}{ICDAR Standard} & \multicolumn{3}{|c|}{DetEval} & \multirow{2}*{FPS} \\
			\cline{2-7}
			~                                     & Recall      & Precision   & F-measure   & Recall      & Precision   & F-measure   & ~          \\
			\hline
			TextBoxes++\cite{TEXTBOX2017AAAI}     & 0.740       & 0.860       & 0.800       & 0.740       & 0.880       & 0.810       & 11.6       \\
			CTPN \cite{CTPN2016ECCV}              & 0.730       & 0.930       & 0.820       & 0.820       & 0.930       & 0.880       & 7.1        \\
			FCRN \cite{FCRN2016CVPR}              & 0.764       & {\bf 0.938} & 0.842       & 0.755       & 0.920       & 0.830       & 0.8        \\
			DeepText \cite{DEEPTEXT2016ARXIV}     & 0.830       & 0.870       & 0.850       & --          & --          & --          & 0.6        \\
			SegLink \cite{SEGLINK2017CVPR}        & --          & --          & --          & 0.830       & 0.877       & 0.853       & 20.6       \\
			Lyu et al. \cite{LYU2018CVPR}         & 0.794       & 0.933       & 0.858       & --          & --          & --          & 10.4       \\
			SSTD \cite{SSTD2017ICCV}              & 0.860       & 0.880       & 0.870       & 0.860       & 0.890       & 0.880       & 7.7        \\
			EAST \cite{EAST2017CVPR}              & 0.828       & 0.926       & 0.874       & --          & --          & --          & 16.8       \\
			CRPN \cite{CRPN2019NEUCO}             & 0.839       & 0.919       & 0.876       & 0.840       & 0.921       & 0.879       & 9.1        \\
			FOTS \cite{EAST2017CVPR}              & --          & --          & 0.882       & --          & --          & 0.883       & 23.9       \\
			{\bf Ours}                            & 0.849       & 0.927       & 0.887       & 0.851       & 0.933       & 0.890       & {\bf 26.5} \\
			\hline
			Lyu et al.$^*$ \cite{LYU2018CVPR}     & 0.844       & 0.920       & 0.880       & --          & --          & --          & 1.0       \\
			TextBoxes++$^*$\cite{TEXTBOX2017AAAI} & 0.840       & 0.910       & 0.880       & 0.860       & 0.920       & 0.890       & 2.3   	   \\
			RRD$^*$ \cite{RRD2018CVPR}            & 0.860       & 0.920       & 0.890       & --          & --          & --          & --   	   \\
			{\bf Ours$^*$}                        & {\bf 0.896} & 0.937       & {\bf 0.916} & {\bf 0.893} & {\bf 0.938} & {\bf 0.915} & 10.5       \\
		\end{tabular}
	\end{center}
	\caption{\textbf{Results on ICDAR 2013 Focused Scene Text.} "*" means multi-scale test. "--" means no report in their papers.}
	\label{tab:icdar2013}
\end{table*}

{\bf ICDAR 2015} \cite{ICDAR2015} benchmark was released during the ICDAR 2015 Robust Reading Competition. It provides 1000 training and 500 testing images which were collected without taking any specific prior attention. It was designed for multi-oriented text detection, so all images are annotated with word-level quadrangles. To evaluate the adaptability of our learned anchor, we still set the input size to 800 pixels. As shown in Table.\ref{tab:icdar2015}, our method achieves an F-measure of 0.833, which also surpasses all of the anchor-based methods \cite{DMPN2017CVPR, SEGLINK2017CVPR, SSTD2017ICCV, RRPN2017ARXIV, TEXTBOX2017AAAI}, including one-stage and two-stage frameworks. Compared with other approaches \cite{EAST2017CVPR, FOTS2018CVPR} which utilize a deep regression network that directly predict text region, our method still keep an absolute lead in running speed. 

\begin{table*}[htb]
	\begin{center}
		\begin{tabular}{c|c|c|c|r} 
			& Recall      & Precision   & F-measure    & FPS        \\
			\hline
			DMPNet \cite{DMPN2017CVPR}         & 0.680       & 0.730       & 0.710        & --         \\
			SegLink \cite{SEGLINK2017CVPR}     & 0.768       & 0.731       & 0.750        & 8.9        \\
			SSTD \cite{SSTD2017ICCV}           & 0.730       & 0.800       & 0.770        & 7.7        \\
			EAST \cite{EAST2017CVPR}           & 0.735       & 0.836       & 0.782        & 13.2       \\
			RRPN \cite{RRPN2017ARXIV}          & 0.770       & 0.840       & 0.800        & 3.0        \\
			TextBoxes++ \cite{TEXTBOX2017AAAI} & 0.707       & {\bf 0.941} & 0.807        & 3.6        \\
			Lyu et al. \cite{LYU2018CVPR}      & 0.767       & 0.872       & 0.817        & 11.6       \\
			RRD \cite{RRD2018CVPR}             & 0.790       & 0.856       & 0.822        & 6.5        \\
			FOTS RT \cite{FOTS2018CVPR}        & 0.798       & 0.856       & 0.828        & 22.6       \\
			CRPN \cite{CRPN2019NEUCO}          & {\bf 0.807} & 0.887       & {\bf 0.845}  & 5.0        \\
			\hline
			\textbf{Ours}                      & 0.786       & 0.887       & 0.833        & {\bf 26.5} \\
		\end{tabular}
	\end{center}
	\caption{\textbf{Results on ICDAR 2015 Incidental Scene Text.} All results of works are reported with single testing scale. "--" means no report in their papers. }
	\label{tab:icdar2015}
\end{table*}

{\bf ICDAR 2017 MLT \cite{ICDAR2017}} is a large scale multi-lingual text dataset, which includes 7200 training, 1800 validation and 9000 testing images with in 9 languages. It was proposed for verifying the generalization ability of each method. Therefore, it is more difficult than previous ICDAR challenges. Due to a larger number of small text instances in this dataset, we enlarge the scale of testing image by 2 times to 1600 pixels and our method achieves 0.655, 0.787, 0.715 in recall, precision and F-measure by using the online evaluation system provided officially, as shown in Table.\ref{tab:icdar2017}. The presented results demonstrate that our method is capable of applying practically in multi-lingual text detection.

\begin{table*}[htb]
	\begin{center}
		\begin{tabular}{c|c|c|c} 
			& Recall      & Precision   & F-measure \\
			\hline
			SARI\_FDU\_RRPN\_v1\textsuperscript{†} & 0.555       & 0.711       & 0.623     \\
			Sensetime OCR\textsuperscript{†}       & {\bf 0.694} & 0.569       & 0.625     \\
			SCUT\_DLVClab1\textsuperscript{†}      & 0.545       & 0.802       & 0.648     \\
			Lyu et al. \cite{LYU2018CVPR}          & 0.556       & {\bf 0.838} & 0.668     \\
			FOTS \cite{FOTS2018CVPR}               & 0.575       & 0.809       & 0.672     \\
			Border \cite{FOTS2018CVPR}             & 0.621       & 0.777       & 0.690     \\
			SPCNET \cite{SPCNET2018ARXIV}          &  0.669      & 0.734       & 0.700     \\
			\hline
			\textbf{Ours}                          & 0.655       & 0.787       & {\bf 0.715} \\
		\end{tabular}
	\end{center}
	\caption{\textbf{Results on ICDAR 2017 Multi-lingual Scene Text Detection.} All results of works are reported with single testing scale. "\textsuperscript{†}" means the result is obtained from the ICDAR 2017 MLT leaderboard.}
	\label{tab:icdar2017}
\end{table*}

%{\bf COCO-Text:} The original images of COCO-Text \cite{COCO2016ARXIV} are harvested from Microsoft COCO \cite{MSCOCO2014ECCV} dataset, and it contains 43686 training images and 20000 images for validation and testing. It is the largest dataset for text detection and recognition in scene images to date. Following the standard practice in object detection, the performances are evaluated by the average precision (AP) at IoU 0.5 and IoU 0.75. All results are calculated by using the online evaluation system provided officially, as shown in Table.\ref{tab:coco}. It is worth noting that no images from COCO-Text are involved in training phase. 
%
%\begin{table}[htb]
%	\begin{center}
%		\begin{tabular}{|c|c|c|}
%			\hline
%			Method              & IoU=0.5     & IoU=0.75    \\
%			\hline\hline
%			FTDN\_SJTU          & 0.320       & 0.079       \\
%			SARI\_FDU\_RRPN     & 0.462       & 0.085       \\
%			SCUT\_DLVClab       & 0.488       & 0.116       \\
%			Text\_Detection\_DL & 0.489       & 0.114       \\
%			\textbf{Ours}       & {\bf 0.524} & {\bf 0.211} \\
%			\hline
%		\end{tabular}
%	\end{center}
%	\caption{\textbf{Results on COCO-Text}. The results of compared methods are grasped from the public COCO-Text leaderboard.}
%	\label{tab:coco}
%\end{table}

%-------------------------------------------------------------------------
\section{Conclusion and Future Work}
In this work, we propose a simple and intuitive method based on RetinaNet for multi-oriented text detection where each location of feature maps associate with only one anchor. The aim of our method is to integrate the learning mechanism from two-stage R-CNN framework into the one-stage detector and utilize the learned anchor to replace the original one into the final predictions. Experimental results on public benchmarks confirm that the proposed method is capable of achieving comparable performance with state-of-the-art methods. Besides, it is a total real-time scene text detector. In the future, we are interested in integrating the detector with a text recognizer to consist an end-to-end text reading system. In addition, we also plan to evaluate it on other detection tasks to prove the universality of our approach. 

%-------------------------------------------------------------------------

{\small
\bibliographystyle{ieee}
\bibliography{egbib}
}

\end{document}